\documentclass[11pt]{article}

\usepackage[final]{acl}

\usepackage{times}
\usepackage{latexsym}
\usepackage{xcolor}

\usepackage{pifont} 
\usepackage{latexsym}
\usepackage{booktabs}
\usepackage{framed}
\usepackage{colortbl}
\usepackage{multirow}
\usepackage{amsmath}
\usepackage{amssymb}
\usepackage{booktabs}
\usepackage{booktabs,array}
\usepackage[table]{xcolor}
\definecolor{GoodBG}{HTML}{E8F5E9}
\definecolor{BadBG}{HTML}{FFEBEE}
\definecolor{GoodText}{HTML}{1B5E20}
\definecolor{BadText}{HTML}{B71C1C}

\usepackage{fvextra}
\DefineVerbatimEnvironment{Verbatim}{Verbatim}{breaklines=true,
 fontsize=\footnotesize
}

\usepackage{booktabs, tabularx, multirow}
\newcolumntype{C}{>{\centering\arraybackslash}X}

\newcolumntype{C}[1]{>{\centering\arraybackslash}m{#1}}
\newcolumntype{L}[1]{>{\raggedright\arraybackslash}m{#1}}

\usepackage{listings}
\usepackage{enumitem}

\usepackage[T1]{fontenc}

\usepackage[utf8]{inputenc}

\usepackage{microtype}

\usepackage{inconsolata}

\usepackage{graphicx}

\newcommand{\aligns}[1]{\textcolor{blue}{#1}}

\title{Beg to Differ: Understanding Reasoning-Answer Misalignment Across Languages}

\author{
 \textbf{Anaelia Ovalle}\textsuperscript{1},
 \textbf{Candace Ross}\textsuperscript{1},
 \textbf{Sebastian Ruder}\textsuperscript{2},
 \textbf{Adina Williams}\textsuperscript{1},
\\
 \textbf{Karen Ullrich}\textsuperscript{1},
 \textbf{Mark Ibrahim}\textsuperscript{1},
 \textbf{Levent Sagun}\textsuperscript{1}
\\[0.5ex]
 \textsuperscript{1}FAIR, Meta \\
 \textsuperscript{2}Meta Superintelligence Labs
}
\begin{document}
\maketitle

\begingroup
\renewcommand\thefootnote{}
\footnotetext{Correspondence: \href{mailto:ccross@meta.com}{ccross@meta.com}}
\addtocounter{footnote}{-1}
\endgroup

\begin{abstract}

\aligns{
}
Large language models demonstrate strong reasoning capabilities through chain-of-thought prompting, but whether this reasoning quality transfers across languages remains underexplored. We introduce a human-validated framework to evaluate whether model-generated reasoning traces logically support their conclusions across languages. Analyzing 65k reasoning traces from GlobalMMLU questions across 6 languages and 6 frontier models, we uncover a critical blind spot: while models achieve high task accuracy, their reasoning can fail to support their conclusions.
Reasoning traces in non-Latin scripts show at least twice as much misalignment between their reasoning and conclusions than those in Latin scripts. We develop an error taxonomy through human annotation to characterize these failures, finding they stem primarily from evidential errors (unsupported claims, ambiguous facts) followed by illogical reasoning steps. Our findings demonstrate that current multilingual evaluation practices provide an incomplete picture of model reasoning capabilities and highlight the need for reasoning-aware evaluation frameworks.
\end{abstract}

\section{Introduction}

\begin{figure}[t]
\centering
    \includegraphics[width=\linewidth]{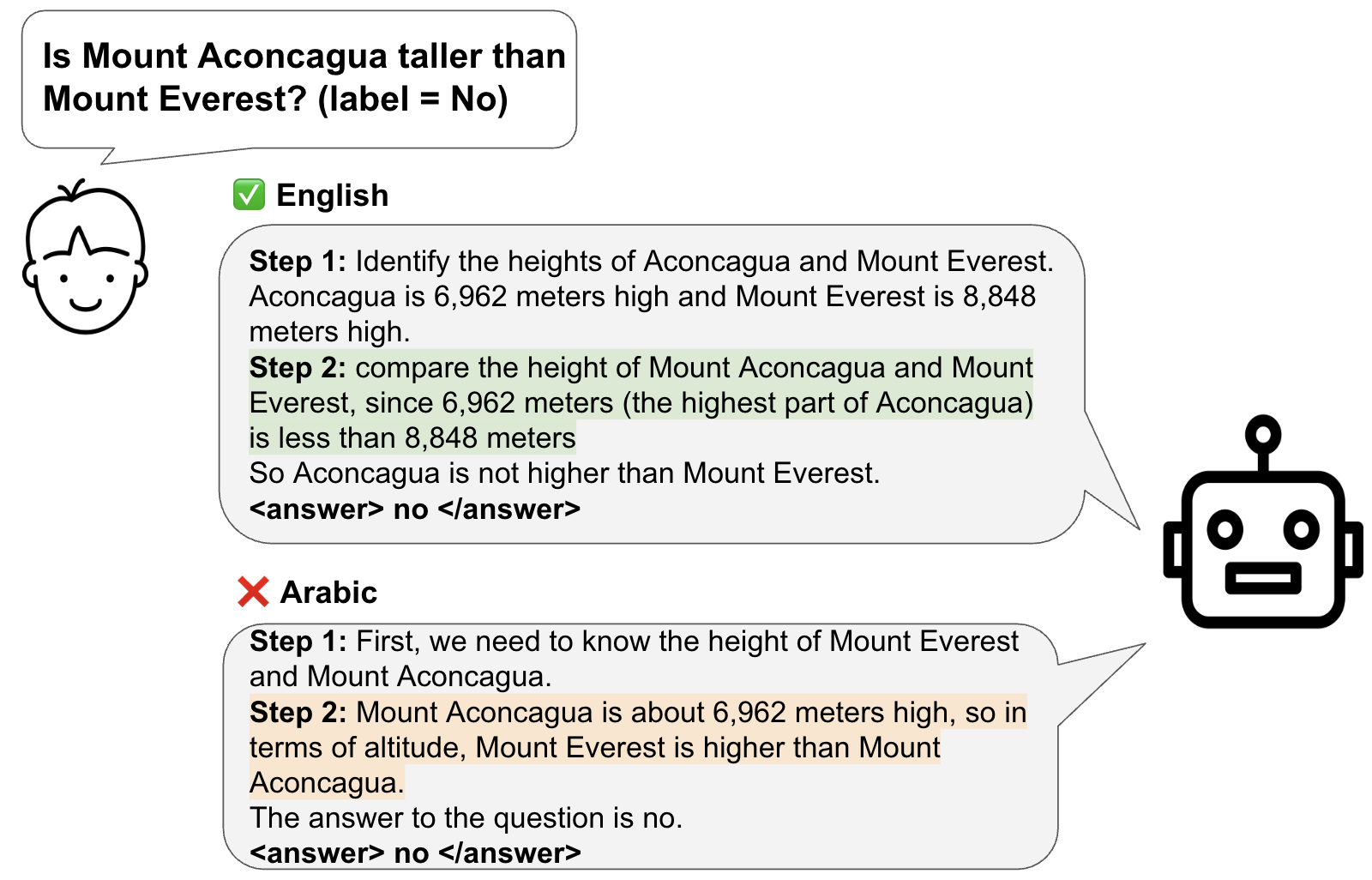}
    \caption{Example of model factuality failure in closed form setting, where the model's final answer is correct both in English and non-English example. however, the model gets the question right for the \textit{wrong reasons} in non-English example. }
    
    \label{fig:mtur-intro}
\end{figure}

Reasoning models have demonstrated remarkable reasoning capabilities through the use of Chain-of-Thought (CoT) prompting, achieving strong performance across several tasks \citep{arcuschin2025chain, lai-nissim-2024-mcot, barua2025longcot}.
However, these reasoning capabilities exhibit significant disparities across languages, with models performing best for high resource languages\citep{shi2022language, qin2023cross, lai-nissim-2024-mcot}.
While multilingual evaluation typically focuses on answer accuracy, we argue this provides an incomplete picture of model capabilities. 
A model might produce correct answers through flawed reasoning (\autoref{fig:mtur-intro}), or conversely, demonstrate sound logical steps while arriving at incorrect conclusions. 
Understanding this distinction is crucial for assessing whether models truly transfer reasoning skills across languages or simply memorize.

To address this, we propose a framework that evaluates whether model-generated reasoning traces logically support their conclusions across languages, measuring crosslingual \textit{reasoning–answer alignment}.
We hypothesize that models exhibit different reasoning processes depending on language resource level and script type, with differences in how they access and use factual knowledge to support their conclusions.
We isolate reasoning traces from their context: evaluators assess whether reasoning steps alone provide sufficient justification to reach the model's conclusion, without seeing the model's final answer.  
Through this human annotation process, we develop an error taxonomy that captures common reasoning failures, which then informs our automated evaluation approach that we validate against the human judgments.
We apply this framework to analyze reasoning traces across five languages with different resource levels and scripts using the Global MMLU dataset. 
We address three key research questions: (RQ1) How does reasoning-answer misalignment vary between high/low resource (or Latin vs. non-Latin script languages)? (RQ2) Can automatic evaluation reliably assess reasoning trace quality compared to human judgment? (RQ3) What are the primary failure modes in multilingual reasoning traces?

Our evaluation framework  reveals a interesting finding:
While answer accuracy predictably dips from high to low-resource languages (80\% -> 70\%), reasoning-answer alignment disproportionately drops over two-fold. 
In other words, models on average, produced LRL reasoning traces that  logically support those conclusions—getting the right answer for the wrong reasons.
We found that lower-resourced languages were able can reach correct conclusions through reasoning that does not logically support those conclusions, getting the right answer for the wrong reasons.
We validate that automated evaluation correlates strongly with human judgment (RQ2), enabling systematic analysis of reasoning failures. 
Critically, this misalignment occurs both when models answer correctly and incorrectly, and varies significantly between Latin and non-Latin script languages (RQ1), indicating systematic issues with reasoning processes rather than simple knowledge gaps.

Our key contributions are: (1) evidence of systematic reasoning-answer misalignment in multilingual settings, with alignment dropping twice as much for non-latin scripts than latin-scripts; (2) human-validated evaluation framework for evaluating reasoning coherence at scale; (3) analysis showing misalignment occurs independently of answer correctness; and (4) characterization of how reasoning failures are distributed across language scripts and subjects.

These findings reveal that answer accuracy provides an incomplete picture of multilingual model capabilities. The reasoning-answer misalignment we identify suggests that models may be learning different problem-solving strategies across languages rather than a learning a uniform and coherent reasoning process. 
Our results indicate that reasoning quality assessment requires separate evaluation from answer correctness in multilingual settings.

\section{Related Works}

\paragraph{Multilingual reasoning evaluation}
Benchmarking multilingual reasoning has largely centered on mathematical domains, with datasets such as MGSM \citep{shi2022language}, MathOctopus \citep{chen2024breaking}, XCOT \citep{chai2025xcot}, PolyMath \citep{wang2025polymath}, and MMATH \citep{luo2025mmath}, where step-by-step solutions can be formally validated.
While mathematical reasoning benefits from clear symbolic representations and verifiable intermediate steps, other domains lack established frameworks for assessing reasoning quality across languages.
Broader multilingual evaluation such as SeaEval \citep{wang2024seaeval}, M3Exam \citep{zhang2023m3exam}, and forms of multilingual value alignment \citep{xu2024exploring, rystrom2025multilingual} evaluate general knowledge and culturally situated reasoning. 
Yet, these works largely equate multilingual reasoning evaluation with performance on reasoning-intensive tasks, judged by whether models reach correct answers, rather than by whether their reasoning processes actually support stated conclusions.

\paragraph{Consistency and faithfulness in LLM reasoning}
Recent work has highlighted persistent issues with the faithfulness of LLM reasoning, particularly for English inputs.
Studies show that chain-of-thought traces often diverge from genuine problem-solving processes \citep{turpin2023language}, with models producing post-hoc rationalizations rather than faithful reasoning explanations \citep{zhang2024language}. These shortcomings manifest both in controlled counterfactual settings \citep{lanham2023measuring} and in real-world scenarios \citep{arcuschin2025chain}.
To address consistency, \citet{wan2024reasoning} propose the RASC framework for evaluating reasoning coherence and guiding early stopping and rationale selection, while \citet{liu2024self} analyze step-level self-contradictions on ambiguous questions. 
In contrast, our focus lies on reasoning trace inference: whether model-generated traces logically support their conclusions in non-ambiguous settings, and crucially, how this differs between non-English languages and language resource level.
Moreover, research that enables automatic prediction of reasoning process failure modes remains understudied.

\begin{table*}[t]
\centering
\small
\setlength{\tabcolsep}{6pt}
\renewcommand{\arraystretch}{1.2}
\begin{tabular}{C{2.5cm} C{3.2cm} L{8cm}@{}}
\toprule
\textbf{Model Answer} & \textbf{Reasoning Trace Support} & \textbf{Description} \\
\midrule
 \cellcolor{GoodBG}{\textcolor{GoodText}{Correct}} 
  & \cellcolor{GoodBG}{\textcolor{GoodText}{Supported}} 
  & Reasoning trace supports the model’s correct answer. \\
 \cellcolor{GoodBG}{\textcolor{GoodText}{Correct}} 
  & \cellcolor{BadBG}{\textcolor{BadText}{Unsupported}} 
  & Reasoning trace does \textbf{not} support the model’s correct answer; the inferred answer differs or is inconclusive. \\
 \cellcolor{BadBG}{\textcolor{BadText}{Incorrect}} 
  & \cellcolor{GoodBG}{\textcolor{GoodText}{Supported}} 
  & Reasoning trace supports the model’s incorrect answer. \\
 \cellcolor{BadBG}{\textcolor{BadText}{Incorrect}} 
  & \cellcolor{BadBG}{\textcolor{BadText}{Unsupported}} 
  & Reasoning trace does \textbf{not} support the model’s incorrect answer; the inferred answer differs or is inconclusive. \\
\bottomrule
\end{tabular}
\caption{Summary of reasoning-trace evaluation cases.}
\label{tab:reasoning_trace_cases}
\end{table*}

\section{Methodology}

We introduce a framework for systematically evaluating reasoning-answer alignment across languages without requiring gold standard reasoning traces. Our approach centers on a key insight: by isolating reasoning content from conclusions, we can assess whether a model's reasoning actually supports its stated answer, which is crucial for developing trustworthy multilingual models.

\subsection{Evaluation Framework}
Given a model that generates reasoning trace $T$ in language $\ell$ concluding with answer $a_{\text{model}}$, we first back-translate non-English responses to English using a top-performing translator for our target languages.\footnote{We do this because LLMs have been shown to perform better when inputs are provided in English \cite{etxaniz-etal-2024-multilingual}.} We then create a truncated trace $T'$ by removing the final answer, allowing us to evaluate what conclusion the reasoning itself supports. Human annotators or automated evaluators assess these truncated traces, selecting from the original multiple-choice options or marking cases as ``inconclusive'' when the reasoning provides insufficient evidence.

We evaluate each response along two dimensions:
(1) \textbf{Correctness}: Whether the model's final answer matches the gold label, and (2) \textbf{Consistency}: Whether the reasoning trace logically supports the model's stated conclusion. These dimensions can be broken down into the cases detailed in \autoref{tab:reasoning_trace_cases}

Our primary metric is the \textbf{Trace Inconsistency Rate} (TIR), the proportion of cases where reasoning fails to support the model's conclusion, regardless of answer correctness.
Additionally, our human evaluation protocol assesses reasoning quality through supplementary measures including logical coherence and information sufficiency.

\subsection{Experimental Setup}

\paragraph{Data and Languages} We conduct our evaluation on GlobalMMLU \citep{singh2024global}, focusing on 7 fact-retrieval centric domains: global facts, high school history (European, US, World), government and politics, and prehistory.

We select languages of varying resource levels and scripts: English, Spanish, Hindi, and Arabic (higher-resourced, HRL), and Ukrainian and Korean (lower-resourced, LRL). 
For script-based analysis, English and Spanish use Latin script (LS), while Hindi, Arabic, Ukrainian, and Korean use non-Latin scripts (NLS).

\begin{table*}[t]
\centering
\small
\begin{tabular}{p{2.5cm}p{8cm}p{4cm}}
\toprule
\textbf{Inconsistency Type} & \textbf{Description} & \textbf{Example Pattern} \\
\midrule
\textbf{Logical Contradiction} & Reasoning steps or claims that conflict with each other, either within the same step or across different steps, resulting contradictory statements that cannot all be simultaneously true.
& The trace first claims option B is correct, then later claims option C is correct without reconciling the contradiction. \\
\midrule
\textbf{Unsupported Claims} & Reasoning contains claims without explicit explanation to supported them. The model may state it relied on "common knowledge" or "standard reasoning", or other implicit assumptions. & "This is obviously true" without explanation \\
\midrule  
\textbf{Multiple Answers} & The reasoning trace concluded with more than one answer option without clear preference.
& The trace concludes with both options A and D. \\

\bottomrule
\end{tabular}
\caption{Sample of our constructed taxonomy of reasoning inconsistencies identified through human annotation and thematic analysis of model-generated reasoning traces. Please see full taxonomy in \autoref{tbl:taxonomyapp}.}
\label{tab:taxonomy}
\end{table*}

\paragraph{Models and Prompting} 
{\raggedright
We evaluate six state-of-the-art instruction-tuned models from two model families:
\texttt{Llama-4-Scout-Instruct},
\texttt{Llama-3.3-70B-Instruct},
\texttt{Qwen3-32B},
\texttt{Qwen3-32B-thinking},
\texttt{Qwen2.5-72B-Instruct}, and
\texttt{Qwen2.5-32B-Instruct}.
\par}\allowbreak All models use chain-of-thought prompting (i.e., ``Let's think step by step'') with structured answer extraction via \texttt{<answer></answer>} tags to ensure consistent outputs across languages and models. More details on the extraction pipeline can be found in the appendix \S\ref{sec:detailspipeline}.

\paragraph{Translation and Backtranslation Quality Control} To enable systematic cross-language comparison, we back-translate all non-English model responses to English using \texttt{Command-R+}, which demonstrates top performance on WMT benchmarks for our target languages \cite{kocmi-etal-2024-findings}. 
We also manually validate translation quality on a small sample of translations, with more details in the appendix \S\ref{sec:detailtranslation}.

\subsection{Evaluation Protocols}
\label{sec:protocols}

\paragraph{Human Evaluation} We design a two-phase human annotation study to systematically evaluate model reasoning quality. In Phase 1, we first conduct a pilot study where 2 expert annotators examine approximately 100 reasoning traces from binary tasks. Taking an inductive approach, we carefully observe patterns in the generated traces, conduct thematic analysis \cite{braun2006using} to identify recurring issues, and iteratively refine our error taxonomy based on these empirical observations.
\autoref{tab:taxonomy} reflects a sample of a constructed taxonomy.

For Phase 2, we scale up our evaluation to GlobalMMLU using stratified sampling across multiple dimensions---language, model, and cultural sensitivity---to ensure broad coverage. We collect 72 annotations total, with 2 annotators per shard to enable reliability analysis. Our annotation protocol asks human evaluators to examine both the original question and the model's reasoning trace $T'$, then complete four key tasks: (1) determine the final answer (selecting from original multiple-choice options or marking ``inconclusive''), (2) assess the logical coherence of the reasoning, (3) evaluate whether the trace contains sufficient information to answer the question, and (4) flag any concerning or problematic aspects of the model's reasoning process.

\paragraph{Automated Evaluation} We replicate the human evaluation protocol using \texttt{Llama-4-Scout-Instruct} as an automated evaluator. The model receives the same truncated traces and applies identical selection criteria.
We employ the evaluator over 66,615 global mmlu examples, keeping only instances where the model returned a valid letter prediction, A-D.

\section{Evaluation Validation}

We validate our evaluation approach through systematic comparison of human and automated assessment across languages and reasoning types.

\subsection{Human Evaluation Results}

Our human evaluation establishes both the reliability of our annotation framework and reveals systematic patterns in reasoning inconsistencies across languages. We achieve substantial inter-annotator agreement on answer selection ($\kappa=0.7$, 80\% agreement), with fair consistency on logical coherence assessment ($\kappa=0.39$, 69.7\% agreement) and moderate-to-strong agreement on information sufficiency ($\kappa=0.53$, 79\% agreement). These results validate that our evaluation protocol captures consistent human judgments about reasoning quality.

Through thematic analysis of the 72 annotated instances, we identify four primary categories of reasoning inconsistencies that emerge systematically across languages and models (Table~\ref{tab:taxonomy}). These failure modes, derived inductively from human observation, reveal how models systematically fail to align their reasoning processes with stated conclusions.

\begin{table}[b]
\centering
\small
\begin{tabular}{lcc}
\toprule
\textbf{Evaluation Dimension} & \textbf{Cohen's $\kappa$} & \textbf{Agreement (\%)} \\
\midrule
Answer Selection & 0.53 & 62.50 \\
Logical Coherence & 0.44 & 70.83 \\
Information Sufficiency & 0.43 & 70.83 \\
\bottomrule
\end{tabular}
\caption{Agreement between automated evaluator and human consensus across evaluation dimensions (n=24 after excluding human disagreements).}
\label{tab:human_auto_agreement}
\end{table}

\begin{table*}[t]
\centering
\footnotesize
\setlength{\tabcolsep}{3pt}
\renewcommand{\arraystretch}{1.08}
\begin{tabular}{l l l l l l l l | l l l l}
\toprule
\textbf{Model Name} & \textbf{Metric} & \textbf{English} & \textbf{Spanish} & \textbf{Hindi} & \textbf{Arabic} & \textbf{Ukrainian} & \textbf{Korean} & \textbf{LS} & \textbf{NLS} & \textbf{HRL} & \textbf{LRL} \\
\midrule
\multirow{2}{*}{\texttt{Llama-4-Scout-Instruct}} 
  & Acc & \textbf{87.06} & 84.33 & 81.31 & 78.71 & 83.95 & 80.33 & 85.7 & 81.11 & 82.93 & 82.2 \\
  & TIR & \textbf{2.61} & 3.99 & 6.87 & 6.12 & 3.71 & 12.57 & 3.3 & 7.25 & 4.86 & 8.0   \\
  \midrule
\multirow{2}{*}{\texttt{Llama-3.3-70B-Instruct}}
  & Acc & \textbf{88.62} & 85.11 & 79.89 & 80.25 & 83.92 & 78.87 & 86.87 & 80.84 & 83.63 & 81.64 \\
  & TIR & \textbf{2.81} & 4.24 & 6.92 & 5.94 & 4.91 & 11.72 & 3.52 & 7.21 & 4.91 & 7.99   \\
\midrule  
\multirow{2}{*}{\texttt{Qwen3-32B}}
  & Acc & \textbf{86.98} & 83.76 & 78.51 & 79.59 & 80.94 & 78.4 & 85.38 & 79.36 & 82.22 & 79.68 \\
  & TIR & \textbf{2.16} & 2.28 & 4.44 & 4.04 & 2.28 & 9.81 & 2.22 & 5.14 & 3.23 & 6.04 \\
\midrule  
\multirow{2}{*}{\texttt{Qwen3-32B-thinking}}
  & Acc & \textbf{88.12} & 85.88 & 81.13 & 81.48 & 83.78 & 81.06 & 87.02 & 81.91 & 84.3 & 82.57 \\
  & TIR & \textbf{0.91} & 1.71 & 3.23 & 3.15 & 2.38 & 8.5 & 1.31 & 4.13 & 2.2 & 5.11   \\
\midrule 
\multirow{2}{*}{\texttt{Qwen2.5-72B-Instruct}}
  & Acc & \textbf{88.77} & 85.84 & 78.29 & 81.48 & 82.78 & 81.8 & 87.31 & 81.05 & 83.67 & 82.39 \\
  & TIR & \textbf{3.11} & 3.5 & 7.96 & 4.51 & 3.75 & 9.63 & 3.3 & 6.16 & 4.74 & 6.09   \\
\midrule 
\multirow{2}{*}{\texttt{Qwen2.5-32B-Instruct}}
  & Acc & \textbf{87.14} & 83.0 & 69.96 & 76.86 & 77.38 & 77.01 & 85.15 & 75.42 & 79.93 & 77.22 \\
  & TIR & 2.96 & \textbf{2.64} & 7.8 & 5.77 & 4.46 & 13.3 & 2.81 & 7.44 & 4.58 & 8.17   \\
\bottomrule
\end{tabular}
\caption{Accuracy (Acc, $\uparrow$) and Trace Inconsistency Rate (TIR, $\downarrow$) across languages and models. 
LS = Latin script, NLS = Not Latin script, HRL = Higher-resourced language, LRL = Lower-resourced language. 
For each model, the highest Accuracy and lowest TIR among individual languages are highlighted in bold.}
\label{tbl:results}
\end{table*}


\subsection{Automated Evaluation Validation}

We validate our automated evaluation by comparing \texttt{Llama-4-Scout-Instruct} performance against human consensus across all evaluation dimensions. To establish reliable human consensus, we aggregate judgments from our two annotators where they agree, excluding the small proportion of cases with human disagreement on answer)

Table~\ref{tab:human_auto_agreement} presents agreement results across our four evaluation dimensions. 
Inferred answer selection shows moderate-to-fairly-strong agreement ($\kappa=0.53$, 62.5\% agreement), while other dimensions of logical coherence and information sufficiency show a slight drop in agreement, perhaps due to an increase in complexity with the labeling task, with $\kappa=0.44$ and $0.43$, respectively. 

Given the considerable agreement on answer selection, we apply our automated evaluator to the GlobalMMLU subset to analyze reasoning-answer consistency patterns at scale, focusing primarily on answer selection where human-automated alignment is strongest.

\begin{table*}[ht]
    \footnotesize
    \setlength{\tabcolsep}{3pt}
    \renewcommand{\arraystretch}{1.08}
    \centering
    \begin{tabular}{l c c c c c c}
        \toprule
        \textbf{Model Name} & \textbf{English} & \textbf{Spanish} & \textbf{Hindi} & \textbf{Arabic} & \textbf{Ukrainian} & \textbf{Korean} \\
        \midrule
        \multirow{1}{*}{Llama-4-Scout-17B-16E-Instruct}
         & \textbf{10.85} & 17.10 & 27.02 & 21.50 & 14.24 & 27.55 \\
        \multirow{1}{*}{Llama-3.3-70B-Instruct}
         & \textbf{10.57} & 14.92 & 23.92 & 21.78 & 20.58 & 22.26 \\
        \multirow{1}{*}{Qwen3-32B}
         & 8.88 & \textbf{8.72} & 16.86 & 14.85 & 9.04 & 18.8 \\
        \multirow{1}{*}{Qwen3-32B-thinking}
         & \textbf{3.40} & 8.09 & 11.59 & 13.25 & 9.00 & 18.0 \\
        \multirow{1}{*}{Qwen2.5-72B-Instruct}
         & \textbf{11.61} & 14.34 & 26.94 & 16.81 & 15.00 & 20.17 \\
        \multirow{1}{*}{Qwen2.5-32B-Instruct}
         & 12.89 & \textbf{10.48} & 22.15 & 21.12 & 15.62 & 28.10 \\
        \bottomrule
    \end{tabular}
    \caption{ Trace Inconsistency Rate (TIR, $\downarrow$) across languages and models, when the model's final answer was \textit{\textbf{wrong}}. For each model, the lowest TIR is bolded.}
    \label{tbl:wrongtir}    
\end{table*}

\section{Results}
\label{sec:results}

As shown in \autoref{tbl:results}, both accuracy and trace inconsistency rates reveal considerable cross-lingual variation, but capture distinct aspects of reasoning quality. Accuracy ranges substantially across languages, where English peaks at 87-89\% while Hindi and Korean drop (70-81\%), and seem consistent with language hierarchies reflective of current literature on language resourcing: English $>$ Spanish $>$ Ukrainian/Arabic $>$ Hindi/Korean \cite{costa2022no}.
However, TIR patterns diverge from accuracy trends, exposing reasoning breakdowns that final answer correctness can obscure.
For instance, while Qwen3-32B maintains reasonable accuracy in Korean (78.4\%), its TIR jumps to 9.81\%, which is over 4$\times$ higher than English (2.16\%). 
Similarly, Qwen2.5-32B-Instruct exhibits 13.3\% Korean TIR despite 77\% accuracy, meaning 1-in-8 responses contain reasoning that contradicts its final answer.
Interestingly though, Qwen3-32B-thinking consistently outperforms its non-thinking version across languages, achieving 0.91\% English TIR versus 2.16\% for the base model, reflecting clear gains of extended reasoning.

Latin script languages (LS) outperform non-Latin script languages (NLS) by 4-6 percentage points in both accuracy and TIR, twice the gap observed between high and low-resource languages!
This pattern seems to hold across models, indicating systematic script-based biases in reasoning quality are stronger than resource availability effects in this study.
We detail explicit examples of reasoning trace failures detected by our evaluating framework in \autoref{sec:error_example_1} and \autoref{sec:error_example_2}.
Furthermore, we provide a subject-level analysis of TIR in \autoref{tbl:subject-inconsistency-answer-right} and \autoref{tbl:subject-level-accuracy}.

\subsection{Trace Inconsistency for Error Diagnosis}

Table~\ref{tbl:wrongtir} shows TIR for cases where the final answer was wrong, where the rates are substantially higher across all languages and models.
Korean, Arabic, and Hindi show the most severe inconsistency when models are incorrect, with some models exceeding 25\% TIR, compared to the 5-13\% rates observed in Table~\ref{tbl:results} overall. 
Even high-resource languages like English and Spanish average 11\% TIR when models are incorrect.
Interestingly, in this setting, these inconsistency rates may reflect their own uncertainty even as they provide incorrect answers, which could serve as canaries for reasoning diagnostics and, we posit, can inform future work on improving model calibration and reasoning reliability.
We provide additional analyses on cases where the model was right, yet still flagged TIR in \autoref{tbl:tir_right}.

\subsection{Analysis of Automated Human-Derived Error Taxonomy}
\begin{figure}[t]
    \includegraphics[width=1\linewidth]{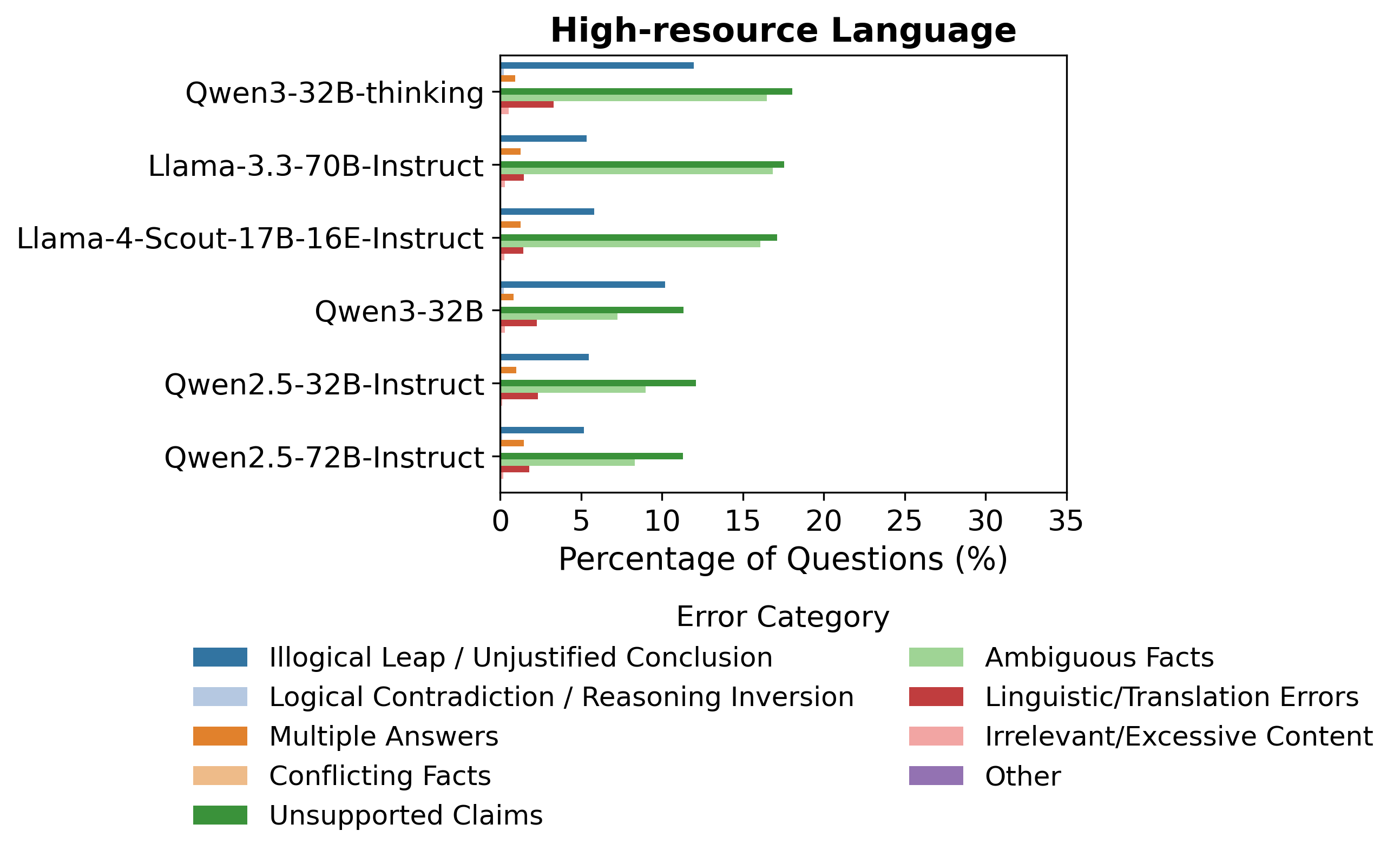}
    \includegraphics[width=1\linewidth]{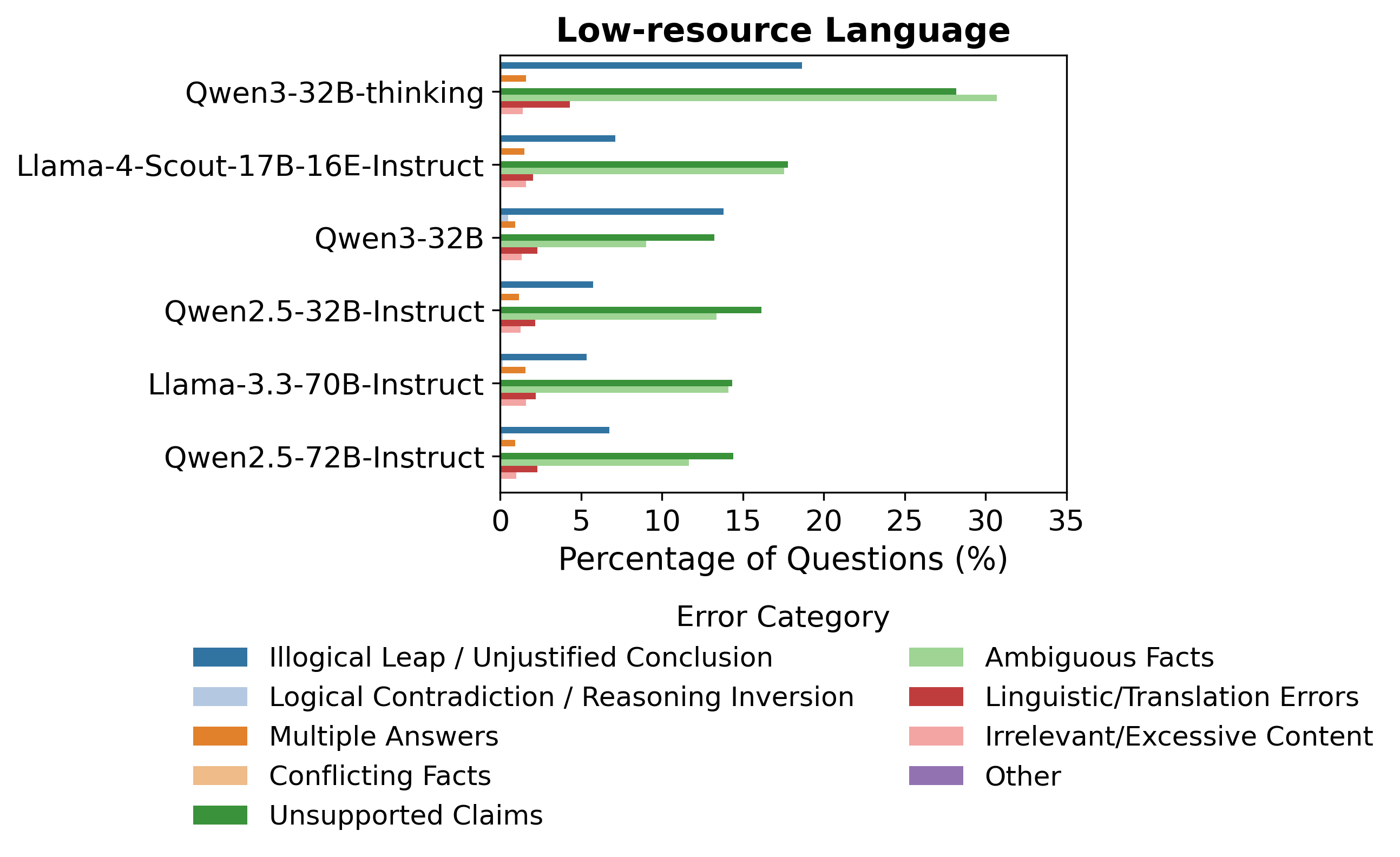}
    \caption{Taxonomy distribution across models. Top is higher-resourced languages, bottom is lower-resourced languages.}
    \label{fig:hrl_error}
\end{figure}

Our automated evaluation framework, employing the human-derived error taxonomy described in \S\ref{sec:protocols}, reveals systematic differences in error patterns across language resource levels (Figure~\ref{fig:hrl_error}). 
Unsupported claims and ambiguous facts constitute the majority of detected errors across all evaluated models (23\% and 24\%respectively), while illogical leaps account for 18\% of flagged instances.
Qwen3-32B-thinking demonstrates the largest resource gap, with 27\% of responses flagged for unsupported claims in LRL settings compared to 18\% in HRL settings.

The prevalence of \textit{evidential errors} (unsupported claims, ambiguous facts) over logical contradictions seems to indicate that models struggle more with grounding their reasoning in verifiable information than with maintaining internal logical consistency. 

Expanding into long-CoT paradigms with thinking models, while they show better consistency overall (as seen in \autoref{sec:results}), the substantial jump in unsupported claims for low-resource languages may hint at potential trade-offs in extended reasoning traces, where while beneficial for complex problems, may drift from well-supported claims when operating with limited linguistic knowledge. Moving forward, we encourage further evaluation on thinking models in this setting.

\section{Conclusion}
In this work we introduced an automatic evaluation framwork to assess reasoning quality through misalignment between trace and answer in a multilingual setting. 
We find that models, even when seemingly similar in accuracy, reflect a varying degree of reasoning trace fidelity. 
While lower-resourced languages are expected to perform worse than higher-resourced languages, accuracy on its own underestimates reasoning trace quality, with languages of non-latin script reflecting twice as many reasoning contradictions than their latin counterparts.
These findings also motivate a closer look at  thinking models, namely, to what extent they can  maintain factual grounding in more resource-constrained language settings.
Moving forward, our evaluation framework can serve to help diagnose reasoning trace errors in order to ultimately improve multilingual reasoning capabilities.

\section*{Limitations}
While our evaluation framework can be employed across other datasets that requiring reasoning, we only test over global mmlu and small sample of boolean comparative questions \cite{arcuschin2025chain}.
We conduct human annotations over a limited number of samples, which can be scaled up in a future iteration.
Furthermore, we only used one judge to assess reasoning traces, where moving forward we can move to using majority voting over 3 judges for robust outputs. 
Evaluation can also take place over a wider set of languages that are even lower-resourced than those initially selected, though our initial set serves for script diversity. 
Translators can also be updated to models like gpt-4o and double-checked by additional native speakers across selected languages.

\section*{Acknowledgments}
We thank Dieuwke Hupkes for helpful discussions on multilingual LM benchmarking.

\bibliography{custom}

\appendix

\section{Example Appendix}
\label{sec:appendix}

\section{Taxonomy}

\begin{table*}[t]
\centering
\footnotesize
\begin{tabular}{p{4cm}|p{6cm}|p{4cm}}
\toprule
\textbf{Error Type} & \textbf{Description} & \textbf{Example Pattern} \\
\midrule
Illogical Leap / Unjustified Conclusion 
& A reasoning step or conclusion is not supported by the stated evidence and may rely on implicit or unstated assumptions. This includes flawed deductive reasoning where the inference from premises to conclusion is logically invalid or incorrect. In other words, the reasoning "jumps" to a conclusion without sufficient logical basis.
& Drawing a conclusion that option A is correct without evidence, or making a logical inference that does not follow from the given premises. \\
\midrule
Logical Contradiction / Reasoning Inversion 
& Reasoning steps or claims that conflict with each other, either within the same step or across different steps, resulting in internal inconsistency. This means the trace contains contradictory statements that cannot all be true simultaneously.
& The trace first claims option B is correct, then later claims option C is correct without reconciling the contradiction.\\
\midrule
Multiple Answers 
& The reasoning trace concluded with more than one answer option without clear preference.
& The trace concludes with both options A and D. \\
\midrule
Conflicting Facts 
& The trace contains facts that contradict each other within the same step or across different steps.
& One step claims a fact that is directly negated in another step.\\
\midrule
Unsupported Claims 
& Claims are made without explicit explanation or data in the reasoning trace. The trace may rely on "common knowledge" or "standard reasoning" without justification.
& Claiming "The product is the best choice because it has superior features," without any objective data or concrete evidence provided in the reasoning trace. \\
\midrule
Ambiguous Facts 
& Claims are underspecified, partial, or lack sufficient detail to support the conclusion clearly.
& Using vague terms like "usually" or "often" without any objective data or concrete evidence provided in the reasoning trace. \\
\midrule
Linguistic/Translation Errors 
& The reasoning trace has issues with coherence, grammar, fluency, or clarity that impede understanding.
& Sentences that are fragmented, poorly translated, or confusing. \\
\midrule
Irrelevant/Excessive Content 
& The trace contains information that is unrelated to the question or unnecessarily verbose beyond what is needed for reasoning.
& Including unrelated background information that does not affect the answer. \\
\midrule
Other 
& A reasoning trace error not already reflected in the taxonomy.
& Any pattern not described above. \\
\bottomrule
\end{tabular}
\caption{Error Taxonomy Table derived from  thematic analysis of reasoning traces during human annotation. }
\label{tbl:taxonomyapp}
\end{table*}

\section{Global MMLU}



\subsection{Pipeline details}
\label{sec:detailspipeline}
All models generate 4096 tokens to respond to each question in global mmlu.
We follow each model's recommended sampling parameters. If none is available, we proceed with nucleus sampling with top-p=0.95.
To account for differences in tokenization across languages, we allot a larger set of tokens for the translator during backtranslation from the original language to english, 8192. Backtranslation is done with temperature set to 0 and top-p set to 1.
Finally, the judge will generates 8192 tokens, and we keep a large window size of 32768 to account for the detailed ai evaluator prompt and the reasoning trace to evaluate. To encourage deterministic output was also set temperature set to 0 and top-p set to 1.

\subsection{Translation Quality Check}
\label{sec:detailtranslation}
We run a small annotation to evaluate the command r plus quality of backtranslation. We employ a native spanish speaker to annotate 30 spanish examples backtranslated to english, and find that only 2/30 examples were marked for errors in translation.

\section*{Reasoning Trace Error Example 1: Unsupported Claims}
\label{sec:error_example_1}

(Model is qwen 2.5 32-b instruct, original language was ukrainian, and sample id was global-facts/test/80.)

\subsection*{Question}
The greatest population growth rate has taken place

\subsection*{Answer Options}
\begin{enumerate}
  \item \textbf{A.} in Africa, which is the poorest region of the world with the lowest overall economic growth.
  \item \textbf{B.} in Asia, which is the poorest region of the world with a steady overall economic growth.
  \item \textbf{C.} in Asia, which is the poorest region of the world with the lowest overall economic growth.
  \item \textbf{D.} in Africa, which is the poorest region of the world with a steady overall economic growth.
\end{enumerate}

\subsection*{Reasoning Trace}
Let's review each answer option:

\begin{itemize}
  \item \textbf{A.} ``In Africa, which is the poorest region in the world with the lowest overall economic growth.''
  
  Reasoning: Africa is the poorest region, but it does not have the lowest economic growth. In fact, African countries have posted significant economic growth rates in recent years.
  
  \item \textbf{B.} ``In Asia, which is the poorest region in the world with steady overall economic growth.''
  
  Reasoning: Asia is not the poorest region of the world, though economic growth has been steady.
  
  \item \textbf{C.} ``In Asia, which is the poorest region in the world with the lowest overall economic growth.''
  
  Reasoning: Factually incorrect; Asia is not the poorest region, nor does it have the lowest economic growth.
  
  \item \textbf{D.} ``In Africa, which is the poorest region in the world with steady overall economic growth.''
  
  Reasoning: Africa matches the description of the poorest region and has experienced both notable and steady economic growth in several countries. It’s also consistently cited as the region with the highest population growth rate.
\end{itemize}

Final answer selected in the reasoning trace: \\
\textbf{D. in Africa, which is the poorest region of the world with a steady overall economic growth.}

\subsection*{Reasoning Trace Error}

\textbf{Primary Error: Unsupported Claims}

\begin{itemize}
  \item \textbf{Description:} The reasoning trace claims that Africa has the highest population growth rate but fails to provide explicit data or evidence to justify that assertion.Without support, the answer may be correct by outside knowledge, but the trace itself is unreliable as reasoning evidence.

\end{itemize}

\subsection*{AI Judge JSON Evaluation}

\begin{Verbatim}
{
  "step_analysis": {
    "A": {"mentioned": "Yes", "supported": "Yes"},
    "B": {"mentioned": "Yes", "supported": "No"},
    "C": {"mentioned": "Yes", "supported": "No"},
    "D": {"mentioned": "Yes", "supported": "Yes"}
  },
  "identified_likely_concluded_answer_or_answers": ["A"], 
  "identified_concluded_explanations": {
    "A": "The trace states that Africa fits the description of the poorest region in the world and has significant economic growth. It is also marked by the highest population growth."
  },
  "identified_errors": ["Unsupported Claims"],
  "error_explanations": {
    "Unsupported Claims": "The trace claims Africa has the highest population growth without providing explicit data or evidence."
  },
  "your_answer": "A"
}
\end{Verbatim}


\section*{Reasoning Trace Error Example 2: Ambiguous Facts and Multiple Answers}
\label{sec:error_example_2}

(Model is llama 3.3 70b instruct, original language was korean, and sample id was global-facts/test/22)

\subsection*{Question}
In 1997, the World Bank found that the fraction of the world’s population living in low and lower-middle income countries—that is, countries with per capita incomes of \$1230 or less—was approximately

\subsection*{Reasoning Trace}
To calculate the percentage of the world's population living in low-income and lower-middle-income countries, i.e., countries with a per capita income of \$1230 or less, according to the World Bank's 1997 report, we need to consider both the population of each country and its per capita income. However, as this detailed data is not provided, we have to rely on general inferences.

Typically, low-income and lower-middle-income countries account for a significant portion of the world's population. Many developing nations fall into this category, and a substantial number of the most populous countries are in this group. For instance, countries like India, China, Indonesia, and Pakistan are in this category, and their populations make up a significant portion of the world's population.

Therefore, it is estimated that the population share of these countries is considerable and may be over 40\%. But to determine the exact percentage, more detailed data and analysis are required.

In conclusion, while it is expected that the percentage of the population living in low-income and lower-middle-income countries is significant, more information is needed to ascertain the exact percentage. If we have to rely on general inferences, we could choose a value close to the median, such as 40\% or 55\%, but more data is necessary.

\subsection*{Reasoning Trace Error}

\textbf{Multiple Errors: Ambiguous Facts and Multiple Answers}
\begin{itemize}
  \item \textbf{Description:} This reasoning trace demonstrates a failure due to ambiguous facts and multiple possible answers. The trace does not provide a clear preference between the two plausible percentages and uses vague language without exact data, which undermines the conclusiveness of the reasoning.
\end{itemize}

\subsection*{AI Judge JSON Response}
\begin{Verbatim}
{
"step_analysis": {
"A": {"mentioned": "No", "supported": "No"},
"B": {"mentioned": "Yes", "supported": "Yes"},
"C": {"mentioned": "Yes", "supported": "Yes"},
"D": {"mentioned": "No", "supported": "No"}
},
"identified_likely_concluded_answer_or_answers": ["B", "C"],
"identified_concluded_explanations": {
"B": "The trace mentions that the percentage could be around 40%.",
"C": "The trace also suggests that the percentage could be around 55%."
},
"identified_errors": ["Multiple Answers", "Ambiguous Facts"],
"error_explanations": {
"Multiple Answers": "The trace concludes with two possible answers, 40% and 55%, without a clear preference.",
"Ambiguous Facts": "The trace uses vague terms like 'considerable' and 'significant portion' without providing exact data."
},
"your_answer": "E"
}
\end{Verbatim}

\newpage
\section{More results}
\label{sec:moreresults}
\begin{table*}[t]
\centering
\footnotesize
\begin{tabular}{lllllll}
\toprule
lang & english & spanish & hindi & arabic & ukrainian & korean \\
model &  &  &  &  &  &  \\
\midrule
Llama-4-Scout-Instruct & 1.38 & 1.56 & 2.24 & 1.96 & 1.69 & 8.91 \\
Llama-3.3-70B-Instruct & 1.81 & 2.37 & 2.64 & 2.05 & 1.91 & 8.9 \\
Qwen3-32B & 1.16 & 1.03 & 1.04 & 1.27 & 0.69 & 7.32 \\
Qwen3-32B-thinking & 0.57 & 0.66 & 1.28 & 0.86 & 1.1 & 6.28 \\
Qwen2.5-72B-Instruct & 2.03 & 1.71 & 2.69 & 1.72 & 1.41 & 7.29 \\
Qwen2.5-32B-Instruct & 1.5 & 1.04 & 1.63 & 1.15 & 1.19 & 8.88 \\
\bottomrule
\end{tabular}
\caption{TIR numbers for when the model's final answer was right.}
\label{tbl:tir_right}
\end{table*}

\begin{table*} [t]
\centering
\footnotesize
\begin{tabular}{lllllll}
\toprule
lang & english & spanish & hindi & arabic & ukrainian & korean \\
model &  &  &  &  &  &  \\
\midrule
meta-llama/Llama-4-Scout-17B-16E-Instruct & 10.85 & 17.1 & 27.02 & 21.5 & 14.24 & 27.55 \\
meta-llama/Llama-3.3-70B-Instruct & 10.57 & 14.92 & 23.92 & 21.78 & 20.58 & 22.26 \\
Qwen/Qwen3-32B & 8.88 & 8.72 & 16.86 & 14.85 & 9.04 & 18.82 \\
Qwen/Qwen3-32B-thinking & 3.4 & 8.09 & 11.59 & 13.25 & 9.0 & 18.02 \\
Qwen/Qwen2.5-72B-Instruct & 11.61 & 14.34 & 26.94 & 16.81 & 15.0 & 20.17 \\
Qwen/Qwen2.5-32B-Instruct & 12.89 & 10.48 & 22.15 & 21.12 & 15.62 & 28.1 \\
\bottomrule
\end{tabular}
\caption{TIR numbers for when the model's final answer was wrong.}
\label{tbl:tir_wrong}
\end{table*}

\begin{table*}[t]
\centering
\footnotesize
\begin{tabular}{llllllll}
\toprule
 & lang & english & spanish & hindi & arabic & ukrainian & korean \\
subject & model &  &  &  &  &  &  \\
\midrule
\multirow[t]{6}{*}{business-ethics} & Llama-4-Scout-17B-16E-Instruct & 79.0 & 76.77 & 76.53 & 74.47 & 75.51 & 72.45 \\
 & Llama-3.3-70B-Instruct & 79.0 & 73.74 & 71.88 & 70.79 & 74.47 & 66.67 \\
 & Qwen3-32B & 84.54 & 86.32 & 72.41 & 77.66 & 73.47 & 78.72 \\
 & Qwen3-32B-thinking & 82.98 & 87.78 & 82.89 & 73.26 & 80.23 & 82.54 \\
 & Qwen2.5-72B-Instruct & 79.0 & 83.67 & 69.07 & 72.63 & 77.78 & 78.67 \\
 & Qwen2.5-32B-Instruct & 81.44 & 76.47 & 67.19 & 76.0 & 77.91 & 80.95 \\
\cline{1-8}
\multirow[t]{6}{*}{global-facts} & Llama-4-Scout-17B-16E-Instruct & 58.0 & 55.56 & 61.29 & 57.45 & 61.0 & 62.37 \\
 & Llama-3.3-70B-Instruct & 65.0 & 62.0 & 59.78 & 59.26 & 66.67 & 55.7 \\
 & Qwen3-32B & 55.0 & 51.0 & 58.76 & 55.56 & 55.0 & 54.0 \\
 & Qwen3-32B-thinking & 56.0 & 53.12 & 54.64 & 49.46 & 54.17 & 59.52 \\
 & Qwen2.5-72B-Instruct & 64.65 & 58.16 & 58.51 & 59.34 & 60.2 & 57.89 \\
 & Qwen2.5-32B-Instruct & 61.0 & 63.83 & 56.25 & 63.41 & 45.16 & 54.29 \\
\cline{1-8}
\multirow[t]{6}{*}{european-history} & Llama-4-Scout-17B-16E-Instruct & 86.06 & 85.8 & 83.95 & 80.95 & 85.62 & 85.43 \\
 & Llama-3.3-70B-Instruct & 84.85 & 84.15 & 84.11 & 81.34 & 85.89 & 85.21 \\
 & Qwen3-32B & 85.37 & 84.15 & 80.37 & 82.42 & 84.15 & 79.27 \\
 & Qwen3-32B-thinking & 86.67 & 83.44 & 82.31 & 85.93 & 85.35 & 82.4 \\
 & Qwen2.5-72B-Instruct & 86.06 & 83.54 & 80.65 & 83.33 & 84.24 & 83.33 \\
 & Qwen2.5-32B-Instruct & 83.03 & 80.75 & 75.94 & 79.61 & 79.11 & 78.74 \\
\cline{1-8}
\multirow[t]{6}{*}{geography} & Llama-4-Scout-17B-16E-Instruct & 91.92 & 90.86 & 86.39 & 80.83 & 86.22 & 85.31 \\
 & Llama-3.3-70B-Instruct & 93.94 & 91.37 & 85.34 & 83.7 & 85.64 & 86.92 \\
 & Qwen3-32B & 90.4 & 87.88 & 84.69 & 82.74 & 84.34 & 82.65 \\
 & Qwen3-32B-thinking & 91.88 & 91.15 & 86.63 & 82.42 & 88.3 & 88.11 \\
 & Qwen2.5-72B-Instruct & 92.93 & 91.37 & 79.35 & 86.91 & 85.2 & 88.35 \\
 & Qwen2.5-32B-Instruct & 90.91 & 85.31 & 73.43 & 77.58 & 81.01 & 87.5 \\
\cline{1-8}
\multirow[t]{6}{*}{government-and-politics} & Llama-4-Scout-17B-16E-Instruct & 96.37 & 89.42 & 88.89 & 86.96 & 91.19 & 91.18 \\
 & Llama-3.3-70B-Instruct & 97.93 & 91.71 & 86.19 & 90.5 & 90.16 & 86.58 \\
 & Qwen3-32B & 98.45 & 91.67 & 83.25 & 89.12 & 89.01 & 86.01 \\
 & Qwen3-32B-thinking & 98.44 & 91.53 & 87.29 & 91.18 & 89.5 & 91.61 \\
 & Qwen2.5-72B-Instruct & 97.93 & 92.63 & 85.95 & 89.58 & 86.53 & 87.4 \\
 & Qwen2.5-32B-Instruct & 97.41 & 92.55 & 74.19 & 82.39 & 79.14 & 84.78 \\
\cline{1-8}
\multirow[t]{6}{*}{us-history} & Llama-4-Scout-17B-16E-Instruct & 92.65 & 92.57 & 89.64 & 87.22 & 92.96 & 84.82 \\
 & Llama-3.3-70B-Instruct & 93.14 & 89.6 & 88.3 & 85.71 & 90.05 & 83.91 \\
 & Qwen3-32B & 90.69 & 87.68 & 83.84 & 82.59 & 88.06 & 79.19 \\
 & Qwen3-32B-thinking & 91.63 & 91.04 & 88.48 & 88.51 & 89.53 & 87.77 \\
 & Qwen2.5-72B-Instruct & 93.14 & 90.95 & 85.05 & 87.24 & 90.59 & 86.36 \\
 & Qwen2.5-32B-Instruct & 91.63 & 87.88 & 72.51 & 83.87 & 85.2 & 79.25 \\
\cline{1-8}
\multirow[t]{6}{*}{world-history} & Llama-4-Scout-17B-16E-Instruct & 89.45 & 86.81 & 87.61 & 82.57 & 87.66 & 85.07 \\
 & Llama-3.3-70B-Instruct & 92.83 & 90.56 & 87.32 & 88.67 & 90.52 & 85.86 \\
 & Qwen3-32B & 88.61 & 85.59 & 83.47 & 86.92 & 84.62 & 85.84 \\
 & Qwen3-32B-thinking & 91.95 & 88.05 & 86.34 & 86.46 & 86.73 & 84.8 \\
 & Qwen2.5-72B-Instruct & 91.14 & 88.14 & 82.97 & 88.21 & 85.41 & 85.88 \\
 & Qwen2.5-32B-Instruct & 89.45 & 88.14 & 77.66 & 84.72 & 84.91 & 77.44 \\
\cline{1-8}
\multirow[t]{6}{*}{logical-fallacies} & Llama-4-Scout-17B-16E-Instruct & 89.57 & 86.5 & 71.78 & 73.62 & 80.37 & 74.38 \\
 & Llama-3.3-70B-Instruct & 87.73 & 80.98 & 65.81 & 73.91 & 78.12 & 69.28 \\
 & Qwen3-32B & 88.89 & 87.26 & 77.78 & 80.75 & 81.25 & 73.58 \\
 & Qwen3-32B-thinking & 90.51 & 86.45 & 73.83 & 80.67 & 82.98 & 70.54 \\
 & Qwen2.5-72B-Instruct & 87.73 & 82.1 & 76.73 & 77.07 & 79.63 & 75.0 \\
 & Qwen2.5-32B-Instruct & 87.73 & 78.62 & 65.55 & 65.28 & 75.0 & 67.62 \\
\cline{1-8}
\multirow[t]{6}{*}{prehistory} & Llama-4-Scout-17B-16E-Instruct & 87.31 & 83.59 & 75.32 & 75.97 & 82.13 & 77.21 \\
 & Llama-3.3-70B-Instruct & 89.81 & 86.65 & 75.42 & 76.08 & 81.9 & 75.0 \\
 & Qwen3-32B & 88.24 & 83.9 & 77.4 & 73.68 & 79.0 & 78.7 \\
 & Qwen3-32B-thinking & 90.09 & 88.14 & 82.06 & 82.14 & 85.71 & 81.54 \\
 & Qwen2.5-72B-Instruct & 91.67 & 87.58 & 76.47 & 81.01 & 84.23 & 82.21 \\
 & Qwen2.5-32B-Instruct & 89.2 & 81.17 & 61.48 & 71.11 & 73.86 & 74.88 \\
\cline{1-8}
\multirow[t]{6}{*}{public-relations} & Llama-4-Scout-17B-16E-Instruct & 71.82 & 71.56 & 70.75 & 66.02 & 73.39 & 64.0 \\
 & Llama-3.3-70B-Instruct & 73.64 & 74.77 & 72.45 & 67.42 & 72.0 & 62.5 \\
 & Qwen3-32B & 73.64 & 72.22 & 61.11 & 68.52 & 66.06 & 60.91 \\
 & Qwen3-32B-thinking & 75.45 & 73.83 & 64.71 & 64.44 & 64.21 & 61.33 \\
 & Qwen2.5-72B-Instruct & 76.36 & 75.7 & 68.0 & 65.35 & 65.74 & 61.02 \\
 & Qwen2.5-32B-Instruct & 74.55 & 68.42 & 54.93 & 64.1 & 61.96 & 58.93 \\
\cline{1-8}
\multirow[t]{6}{*}{sociology} & Llama-4-Scout-17B-16E-Instruct & 90.05 & 85.0 & 86.46 & 82.56 & 87.31 & 83.16 \\
 & Llama-3.3-70B-Instruct & 91.54 & 86.07 & 84.78 & 80.81 & 86.29 & 85.12 \\
 & Qwen3-32B & 89.05 & 84.08 & 79.29 & 80.6 & 83.42 & 81.82 \\
 & Qwen3-32B-thinking & 88.0 & 87.24 & 81.77 & 84.75 & 87.29 & 81.33 \\
 & Qwen2.5-72B-Instruct & 90.55 & 88.38 & 79.49 & 80.3 & 87.56 & 86.76 \\
 & Qwen2.5-32B-Instruct & 88.56 & 83.42 & 72.33 & 78.02 & 78.76 & 78.57 \\
\cline{1-8}
\bottomrule
\end{tabular}
\caption{subject-level accuracy on global mmlu}
\label{tbl:subject-level-accuracy}
\end{table*}

\begin{table*}
\centering
\footnotesize
    \begin{tabular}{llllllll}
\toprule
 & lang & english & spanish & hindi & arabic & ukrainian & korean \\
subject & model &  &  &  &  &  &  \\
\midrule
\multirow[t]{6}{*}{business-ethics} & Llama-4-Scout-17B-16E-Instruct & 3.8 & 0.0 & 0.0 & 0.0 & 1.35 & 1.41 \\
 & Llama-3.3-70B-Instruct & 2.53 & 1.37 & 5.8 & 1.59 & 0.0 & 3.85 \\
 & Qwen3-32B & 1.22 & 0.0 & 0.0 & 0.0 & 0.0 & 0.0 \\
 & Qwen3-32B-thinking & 0.0 & 1.27 & 0.0 & 0.0 & 0.0 & 0.0 \\
 & Qwen2.5-72B-Instruct & 2.53 & 2.44 & 2.99 & 1.45 & 1.3 & 1.69 \\
 & Qwen2.5-32B-Instruct & 1.27 & 0.0 & 2.33 & 0.0 & 1.49 & 3.92 \\
\cline{1-8}
\multirow[t]{6}{*}{global-facts} & Llama-4-Scout-17B-16E-Instruct & 3.45 & 5.45 & 5.26 & 5.56 & 6.56 & 10.34 \\
 & Llama-3.3-70B-Instruct & 10.77 & 12.9 & 9.09 & 9.38 & 5.36 & 2.27 \\
 & Qwen3-32B & 0.0 & 3.92 & 0.0 & 0.0 & 0.0 & 0.0 \\
 & Qwen3-32B-thinking & 0.0 & 1.96 & 0.0 & 0.0 & 0.0 & 2.0 \\
 & Qwen2.5-72B-Instruct & 3.12 & 1.75 & 1.82 & 3.7 & 1.69 & 3.03 \\
 & Qwen2.5-32B-Instruct & 1.64 & 3.33 & 0.0 & 7.69 & 0.0 & 10.53 \\
\cline{1-8}
\multirow[t]{6}{*}{european-history} & Llama-4-Scout-17B-16E-Instruct & 2.11 & 1.44 & 2.21 & 2.52 & 0.73 & 2.33 \\
 & Llama-3.3-70B-Instruct & 0.71 & 0.0 & 2.36 & 0.92 & 4.29 & 0.83 \\
 & Qwen3-32B & 0.71 & 0.72 & 2.29 & 1.47 & 0.72 & 0.0 \\
 & Qwen3-32B-thinking & 0.0 & 0.0 & 2.48 & 0.0 & 0.75 & 0.97 \\
 & Qwen2.5-72B-Instruct & 2.82 & 0.0 & 2.4 & 0.74 & 0.72 & 1.05 \\
 & Qwen2.5-32B-Instruct & 1.46 & 0.77 & 1.98 & 0.83 & 1.6 & 2.0 \\
\cline{1-8}
\multirow[t]{6}{*}{geography} & Llama-4-Scout-17B-16E-Instruct & 1.1 & 1.12 & 3.03 & 1.28 & 2.37 & 3.97 \\
 & Llama-3.3-70B-Instruct & 1.61 & 2.78 & 3.68 & 2.6 & 2.4 & 1.77 \\
 & Qwen3-32B & 1.68 & 0.57 & 1.2 & 0.0 & 1.2 & 1.85 \\
 & Qwen3-32B-thinking & 1.1 & 0.57 & 1.85 & 1.47 & 1.81 & 1.59 \\
 & Qwen2.5-72B-Instruct & 1.09 & 1.11 & 1.37 & 1.2 & 2.4 & 1.1 \\
 & Qwen2.5-32B-Instruct & 1.67 & 1.99 & 0.0 & 2.34 & 0.69 & 1.19 \\
\cline{1-8}
\multirow[t]{6}{*}{government-and-politics} & Llama-4-Scout-17B-16E-Instruct & 0.0 & 0.0 & 1.19 & 0.62 & 0.57 & 0.65 \\
 & Llama-3.3-70B-Instruct & 0.0 & 1.13 & 1.92 & 3.7 & 0.57 & 0.78 \\
 & Qwen3-32B & 0.0 & 0.0 & 1.26 & 0.58 & 0.0 & 0.0 \\
 & Qwen3-32B-thinking & 0.0 & 0.58 & 1.27 & 0.65 & 0.0 & 0.0 \\
 & Qwen2.5-72B-Instruct & 0.0 & 0.57 & 3.14 & 1.74 & 1.2 & 0.0 \\
 & Qwen2.5-32B-Instruct & 0.53 & 1.15 & 1.74 & 0.0 & 1.35 & 0.85 \\
\cline{1-8}
\multirow[t]{6}{*}{us-history} & Llama-4-Scout-17B-16E-Instruct & 1.59 & 2.14 & 4.05 & 1.91 & 1.62 & 2.47 \\
 & Llama-3.3-70B-Instruct & 2.11 & 0.55 & 2.41 & 2.0 & 1.66 & 3.42 \\
 & Qwen3-32B & 0.54 & 2.25 & 0.0 & 0.6 & 0.0 & 0.64 \\
 & Qwen3-32B-thinking & 0.0 & 0.0 & 0.59 & 0.65 & 0.58 & 0.0 \\
 & Qwen2.5-72B-Instruct & 1.58 & 0.0 & 1.82 & 2.34 & 1.64 & 0.88 \\
 & Qwen2.5-32B-Instruct & 1.08 & 0.57 & 3.23 & 0.0 & 1.2 & 0.79 \\
\cline{1-8}
\multirow[t]{6}{*}{world-history} & Llama-4-Scout-17B-16E-Instruct & 1.42 & 0.98 & 1.01 & 1.11 & 0.97 & 50.0 \\
 & Llama-3.3-70B-Instruct & 0.91 & 1.9 & 0.54 & 1.11 & 0.48 & 51.18 \\
 & Qwen3-32B & 0.48 & 0.5 & 1.02 & 1.46 & 0.0 & 50.0 \\
 & Qwen3-32B-thinking & 0.0 & 0.0 & 1.13 & 1.2 & 0.0 & 44.14 \\
 & Qwen2.5-72B-Instruct & 0.93 & 0.0 & 1.05 & 0.5 & 0.5 & 44.74 \\
 & Qwen2.5-32B-Instruct & 0.0 & 0.0 & 0.0 & 0.52 & 0.51 & 47.68 \\
\cline{1-8}
\multirow[t]{6}{*}{logical-fallacies} & Llama-4-Scout-17B-16E-Instruct & 0.68 & 2.13 & 1.71 & 3.33 & 0.76 & 1.68 \\
 & Llama-3.3-70B-Instruct & 1.4 & 2.27 & 2.94 & 0.84 & 0.8 & 1.89 \\
 & Qwen3-32B & 4.17 & 0.0 & 1.59 & 1.54 & 0.0 & 0.85 \\
 & Qwen3-32B-thinking & 2.1 & 0.0 & 0.0 & 0.0 & 0.85 & 2.2 \\
 & Qwen2.5-72B-Instruct & 2.1 & 2.26 & 2.46 & 0.0 & 0.78 & 2.22 \\
 & Qwen2.5-32B-Instruct & 2.1 & 1.6 & 1.28 & 2.13 & 1.8 & 0.0 \\
\cline{1-8}
\multirow[t]{6}{*}{prehistory} & Llama-4-Scout-17B-16E-Instruct & 1.06 & 2.59 & 2.59 & 1.28 & 1.15 & 1.76 \\
 & Llama-3.3-70B-Instruct & 1.37 & 2.51 & 1.32 & 0.87 & 2.33 & 1.06 \\
 & Qwen3-32B & 1.05 & 1.48 & 1.2 & 1.26 & 0.79 & 1.96 \\
 & Qwen3-32B-thinking & 1.03 & 0.36 & 0.81 & 0.87 & 1.14 & 0.47 \\
 & Qwen2.5-72B-Instruct & 2.02 & 3.55 & 2.99 & 2.34 & 0.75 & 1.17 \\
 & Qwen2.5-32B-Instruct & 1.04 & 2.0 & 2.0 & 1.56 & 0.0 & 1.97 \\
\cline{1-8}
\multirow[t]{6}{*}{public-relations} & Llama-4-Scout-17B-16E-Instruct & 2.53 & 1.28 & 2.67 & 5.88 & 7.5 & 9.38 \\
 & Llama-3.3-70B-Instruct & 6.17 & 8.75 & 8.45 & 3.33 & 5.56 & 4.44 \\
 & Qwen3-32B & 2.47 & 3.85 & 1.52 & 6.76 & 5.56 & 0.0 \\
 & Qwen3-32B-thinking & 2.41 & 5.06 & 6.06 & 3.45 & 6.56 & 4.35 \\
 & Qwen2.5-72B-Instruct & 8.33 & 6.17 & 5.88 & 3.03 & 2.82 & 0.0 \\
 & Qwen2.5-32B-Instruct & 6.1 & 0.0 & 7.69 & 4.0 & 1.75 & 12.12 \\
\cline{1-8}
\multirow[t]{6}{*}{sociology} & Llama-4-Scout-17B-16E-Instruct & 1.1 & 1.18 & 1.81 & 2.48 & 1.16 & 3.16 \\
 & Llama-3.3-70B-Instruct & 1.09 & 1.16 & 0.64 & 2.5 & 1.18 & 4.9 \\
 & Qwen3-32B & 1.12 & 0.59 & 0.64 & 1.85 & 1.2 & 1.85 \\
 & Qwen3-32B-thinking & 0.0 & 1.17 & 1.27 & 1.33 & 2.53 & 2.46 \\
 & Qwen2.5-72B-Instruct & 2.75 & 2.86 & 5.16 & 3.14 & 2.84 & 0.85 \\
 & Qwen2.5-32B-Instruct & 2.81 & 0.6 & 0.87 & 0.7 & 3.29 & 2.48 \\
\cline{1-8}
\bottomrule
\end{tabular}
\caption{per subject level inconsistency, when the final answer was right.}
\label{tbl:subject-inconsistency-answer-right}
\end{table*}

\section{AI Judge Evaluator Prompt}
\label{sec:aijudgeprompt}

The complete prompt used for reasoning trace analysis is provided below:

\begin{Verbatim}
## TASK
Analyze the **Question**, **Options**, and **Reasoning Trace** provided.  
Your goal is to infer the answer option that the reasoning trace most likely 
supported or concluded -- even if the reasoning trace support was weak, confused, 
ambiguous, or based on assumptions.  
You do **not** mark errors during this process; error identification is done 
separately after the answer inference.

## CONSTRAINTS
- Use only the information present in the **Question**, **Options**, and 
  **Reasoning Trace**.  
- Do **not** use external knowledge (e.g., facts or information not explicitly 
  stated in the question, options, or reasoning trace), assumptions, or 
  supplement/fix/correct the reasoning trace, even if it appears confused.  
- Do **not** generate commentary, explanation, or any content outside the 
  required output format.  
- If the trace does not clearly lean toward any option, or ambiguously supports 
  multiple options, select E (Unknown).  

## ERROR TAXONOMY
Error identification is performed **after** answer selection. When marking 
errors, consider these error types. Multiple error types may be assigned if 
applicable.

- **Illogical Leap / Unjustified Conclusion**  
  A reasoning step or conclusion is not supported by the stated evidence and may 
  rely on implicit or unstated assumptions. This includes flawed deductive 
  reasoning where the inference from premises to conclusion is logically invalid 
  or incorrect. In other words, the reasoning "jumps" to a conclusion without 
  sufficient logical basis.  
  *Example:* Drawing a conclusion that option A is correct without evidence, or 
  making a logical inference that does not follow from the given premises.

- **Logical Contradiction / Reasoning Inversion**  
  Reasoning steps or claims that conflict with each other, either within the same 
  step or across different steps, resulting in internal inconsistency.  
  This means the trace contains contradictory statements that cannot all be true 
  simultaneously.
  *Example:* The trace first claims option B is correct, then later claims 
  option C is correct without reconciling the contradiction.

- **Multiple Answers**  
  The reasoning trace concluded with more than one answer option without clear 
  preference.  
  *Example:* The trace concludes with both options A and D.

- **Conflicting Facts**  
  The trace contains facts that contradict each other within the same step or 
  across different steps.  
  *Example:* One step claims a fact that is directly negated in another step.

- **Unsupported Claims**  
  Claims are made without explicit explanation or data in the reasoning trace. 
  The trace may rely on "common knowledge" or "standard reasoning" without 
  justification.  
  *Example:* Claiming "The product is the best choice because it has superior 
  features," without any objective data or concrete evidence provided in the 
  reasoning trace.

- **Ambiguous Facts**  
  Claims are underspecified, partial, or lack sufficient detail to support the 
  conclusion clearly.  
  *Example:* Using vague terms like "usually" or "often" without any objective 
  data or concrete evidence provided in the reasoning trace.

- **Linguistic/Translation Errors**  
  The reasoning trace has issues with coherence, grammar, fluency, or clarity 
  that impede understanding.  
  *Example:* Sentences that are fragmented, poorly translated, or confusing.

- **Irrelevant/Excessive Content**  
  The trace contains information that is unrelated to the question or 
  unnecessarily verbose beyond what is needed for reasoning.  
  *Example:* Including unrelated background information that does not affect 
  the answer.

- **Other**  
  A reasoning trace error not already reflected in the taxonomy.

## INSTRUCTIONS
1. Identify which answer option(s) the reasoning trace most likely concluded 
   with in the output JSON under `"identified_likely_concluded_answer_or_answers"`.  
2. For each option, mark if they are mentioned and then if were favored in the 
   output JSON under `"step_analysis"`.
3. If you can unambiguously infer from the trace the concluded answer, you 
   should select that option in the output JSON under 
   `"identified_likely_concluded_answer_or_answers"`.
4. If you cannot infer from trace the concluded answer, or the trace could have 
   ambiguously concluded with multiple options, you should select those options 
   in the output JSON under `"identified_likely_concluded_answer_or_answers"`, 
   then your answer should be E (Unknown) under `"your_answer"`.
5. After your answer, provide a brief explaation in the output JSON under 
   `"identified_concluded_explanations"`.
6. After your selection, mark which errors, if any, are present in the reasoning 
   trace using the provided error taxonomy in the output JSON under 
   `"identified_errors"`.  
7. For each identified error, provide a brief explanation in the output JSON 
   under `"error_explanations"`.  
8. Do **not** generate any content outside the required JSON output.

## QUESTION
{question}

## OPTIONS
{options}
E. Unknown.

## REASONING TRACE
{reasoning}

## OUTPUT FORMAT
Respond **only** with valid JSON in the following format.  
Do not add explanations, commentary, or any extra content outside the JSON.

```json
{
  "step_analysis": {
    "A": {"mentioned": "Yes" | "No", "supported": "Yes" | "No"},
    "B": {"mentioned": "Yes" | "No", "supported": "Yes" | "No"},
    "C": {"mentioned": "Yes" | "No", "supported": "Yes" | "No"},
    "D": {"mentioned": "Yes" | "No", "supported": "Yes" | "No"}
  },
  "identified_likely_concluded_answer_or_answers": ["A" | "B" | "C" | "D" | "None"],
  "likely_concluded_explanations": {},
  "identified_errors": [],
  "error_explanations": {},
  "your_answer": "A" | "B" | "C" | "D" | "E"
}
```

## FEW-SHOT EXAMPLES
### Example 1: Single Identified Error
### QUESTION:
What is the largest planet in the solar system?
### OPTIONS:
A: Earth
B: Mars
C: Jupiter
D: Venus
E: Unknown
### REASONING TRACE:
The trace states: "I think the answer is B because Mars is the biggest planet."
### OUTPUT:

```json
{
  "step_analysis": {
    "A": {"mentioned": "No", "supported": "No"},
    "B": {"mentioned": "Yes", "supported": "Yes"},
    "C": {"mentioned": "No", "supported": "No"},
    "D": {"mentioned": "No", "supported": "No"}
  },
  "identified_likely_concluded_answer_or_answers": ["B"],
  "identified_errors": ["Unsupported Claims"],
  "likely_concluded_explanations": {
    "B": "The trace claims Mars is the biggest planet."
  },
  "error_explanations": {
    "Unsupported Claims": "The trace claims Mars is the biggest planet without evidence."
  },
  "your_answer": "B"
}
```

### Example 2: Multiple Identified Errors
### QUESTION:
Which animal is known as the "King of the Jungle"?
### OPTIONS:
A: Lion
B: Tiger
C: Elephant
D: Bear
E: Unknown
### REASONING TRACE:
The trace states: "The answer is Lions because they live in the jungle and are 
the strongest animals. Also, tigers are known as the kings of the jungle."      
### OUTPUT:

```json
{
  "step_analysis": {
    "A": {"mentioned": "Yes", "supported": "Yes"},
    "B": {"mentioned": "Yes", "supported": "Yes"},
    "C": {"mentioned": "No", "supported": "No"},
    "D": {"mentioned": "No", "supported": "No"}
  },
  "identified_likely_concluded_answer_or_answers": ["A", "B"],
  "likely_concluded_explanations": {
    "A": "The trace claims Lions are the kings because they are the strongest and live in the jungle.",
    "B": "The trace also said tigers are known as the kings of the jungle."
  },        
  "identified_errors": ["Multiple Options Supported", "Conflicting Facts"],
  "error_explanations": {
    "Multiple Options Supported": "The trace ambiguously supports both options A and B.",
    "Conflicting Facts": "The trace claims both lions and tigers are the 'King of the Jungle', which is contradictory."
  },
  "your_answer": "E"
}
```

### Example 3: No error
### QUESTION:
What is the largest planet in the solar system?
### OPTIONS:
A: Earth
B: Mars
C: Jupiter
D: Venus
E: Unknown
### REASONING TRACE:
The trace states: "A isn't it because Earth is 5ft. B is not the largest because 
its 2ft. D isn't the largest because its 1ft."
### OUTPUT:

```json
{
  "step_analysis": {
    "A": {"mentioned": "Yes", "supported": "No"},
    "B": {"mentioned": "Yes", "supported": "No"},
    "C": {"mentioned": "No", "supported": "No"},
    "D": {"mentioned": "Yes", "supported": "No"}
  },
  "identified_likely_concluded_answer_or_answers": ["C"],
  "likely_concluded_explanations": {
    "C": "The trace concluded C because it explicitly eliminated all other possible options."
  },         
  "identified_errors": [""],
  "error_explanations": {},
  "your_answer": "C"
}
```
\end{Verbatim}

\end{document}